\documentclass[10pt,twocolumn,letterpaper]{article}

\usepackage[pagenumbers]{cvpr} 

\usepackage{graphicx}
\usepackage{amsmath}
\usepackage{amssymb}
\usepackage{booktabs}
\usepackage{rotating}
\usepackage{multirow}

\usepackage[pagebackref,breaklinks,colorlinks]{hyperref}

\usepackage[capitalize]{cleveref}
\crefname{section}{Sec.}{Secs.}
\Crefname{section}{Section}{Sections}
\Crefname{table}{Table}{Tables}
\crefname{table}{Tab.}{Tabs.}

\def\cvprPaperID{*****} 
\def\confName{CVPR}
\def\confYear{2023}

\begin{document}

\title{Generating Context-Aware Natural Answers for Questions in 3D Scenes}

\author{
    Mohammed Munzer Dwedari \quad \quad Matthias Niessner \quad \quad Dave Zhenyu Chen \vspace{0.6em} \\
    {\normalsize
        Technical University of Munich
    }
}

\def\eg{\emph{e.g.}}
\def\Eg{\emph{E.g.}}
\def\etal{\emph{et al.}}

\newcommand{\mypara}[1]{\noindent \textbf{#1}}

\maketitle

\begin{abstract}
3D question answering is a young field in 3D vision-language that is yet to be explored. 
Previous methods are limited to a pre-defined answer space and cannot generate answers naturally.
In this work, we pivot the question answering task to a sequence generation task to generate free-form natural answers for questions in 3D scenes (Gen3DQA).
To this end, we optimize our model directly on the language rewards to secure the global sentence semantics.
Here, we also adapt a pragmatic language understanding reward to further improve the sentence quality. 
Our method sets a new SOTA on the ScanQA benchmark (CIDEr score \textbf{72.22/66.57} on the test sets). The project code can be found at: \url{https://github.com/MunzerDw/Gen3DQA.git}.
\end{abstract}

\section{Introduction}

Visual question answering is a fundamental task in vision-language understanding~\cite{antol2015vqa, agrawal2016vqa, shahMYP2019KVQA, srivastava2020vqasurvey, ye20223dvqa}. Unlike dense captioning~\cite{johnson2016densecap, yang2017dense, kim2019dense, chen2021scan2cap} or visual grounding~\cite{hu2016natural, hu2016segmentation, kazemzadeh2014referitgame, yu2016modeling, yu2017joint, deng2018visual, chen2020scanrefer, chen2022unit3d}, 
question answering requires the intelligent system to understand the joint context of the question (language) and scene (vision) to interact with the environment by providing answers. While visual question answering on images has been extensively researched, question answering on 3D scenes is yet to be explored. 

The seminal work such as ScanQA~\cite{azuma2022scanqa} relies on a two-branch architecture for encoding both modalities of the input (point cloud and question) before fusing them into a joint vector representation. 
Then, an answer is predicted among a predefined answer space using such multimodal feature. 
Along with the answer classification, a bounding box is predicted to localize the object referred to in the question. 
Another competitive method, CLIP-guided~\cite{parelli2023clipguided}, transfers 2D prior knowledge to the 3D domain via a contrastive learning scheme using CLIP features~\cite{radford2021learning}. However, aforementioned baseline methods are limited to a predefined answer space, which consequently hinders the capability of interacting with human users.

To tackle this challenge, we propose a transformer-based architecture to generate, rather than predict, free-from answers for questions in 3D environments. Using reinforcement learning, we directly train our model with a language reward to secure the global semantics of the generated sentences, as shown in Figure~\ref{fig:teaser}. To this end, we utilize the policy gradient method~\cite{rennie2017selfcritical} to approximate sampled gradients through our end-to-end architecture. To further ensure the correctness of the generated answers, we introduce an additional helper reward that encourages the model to reversely reconstruct the questions from the respective answers. 
Our method outperforms the state-of-the-art methods for the image captioning metrics on the ScanQA~\cite{azuma2022scanqa} benchmark.
We summarize our contributions as follows:

\begin{figure}
    \centering
    \includegraphics[width=1.0\linewidth]{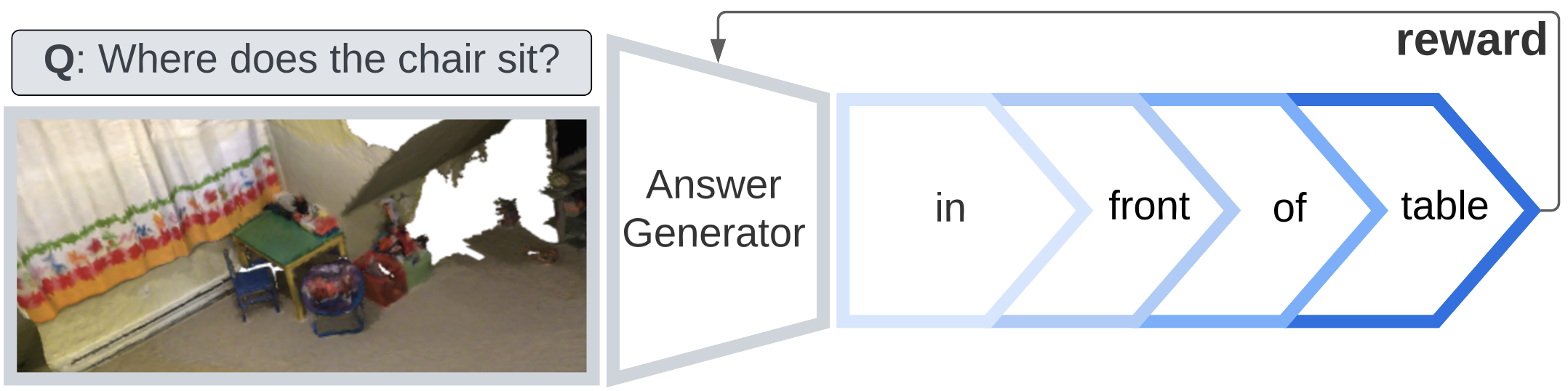}
    \caption{
    We propose Gen3DQA, an end-to-end transformer-based architecture for generating natural answers for questions in 3D scenes. Our method directly optimizes the global semantics of the generated sentences via the language rewards.
    }
    \label{fig:teaser}
\end{figure}

\begin{itemize}
  \item We propose an end-to-end transformer-based architecture for the task of 3D visual question answering, which deals with ambiguous contexts and generates free-form natural answers.
  \item We present a reinforcement-learning-based training objective that directly optimizes the global semantics of the generated sentences. We also incorporate a pragmatic helper reward that encourages correct reconstruction of the question from the generated answer, which further improves answer quality.
  \item We conduct extensive experiments and ablation studies to show the effectiveness of our method. Our method sets a new SOTA performance on the ScanQA benchmark~\cite{azuma2022scanqa} (CIDEr score \textbf{72.22/66.57} on the test sets).
\end{itemize}

\section{Related Work}


\mypara{3D Vision-Language.} One of the first works to combine 3D scenes and natural language is the ScanRefer ~\cite{chen2020scanrefer} benchmark, which introduces the task of visual grounding in ScanNet~\cite{dai2017scannet} scenes. The ScanRefer~\cite{chen2020scanrefer} dataset has more object categories than originally set in ScanNet~\cite{dai2017scannet}. The authors design a two branch model where one branch encodes the scene with a PointNet++~\cite{qi2017pointnet++} backbone and the other the reference sentence with a GRU~\cite{chung2014empirical}. A fusion module takes in both encoded modalities to predict the final target object. Shortly after, the reverse task of dense captioning in 3D scenes is introduced~\cite{chen2021scan2cap}. The same backbone is used in addition to a graph module and attention mechanism to predict the bounding boxes and generate their descriptions. D3Net~\cite{chen2022d3net} combines both tasks into a speaker-listener architecture, where the speaker takes in predicted object proposals from a detector backbone and generates a caption sentence for each one. The generated captions are passed to the listener, which grounds the target objects. The method employs the REINFORCE~\cite{Williams2004SimpleSG} algorithm for sequence generation~\cite{rennie2017selfcritical} to train both modules jointly.

\mypara{Visual Question Answering.} Question answering~\cite{srivastava2020vqasurvey} on images has been extensively researched, where most models approach the problem as a classification task~\cite{agrawal2016vqa, yang2016stacked, teney2017tips, yi2019neuralsymbolic, jiang2018pythia}. In addition, several works focus on VQA with video as input~\cite{Sadhu2021VideoQA, xiao2021video, invariant2022video, zhong2022Video}. Today, large transformer based pretrained models have shown the ability to achieve superior performance on the VQA task~\cite{tan2019lxmert, lu2019vilbert, su2020vlbert, srivastava2020vqasurvey, wang2022beit, bao2022vlmo, chen2022pali, driess2023palme}. However, in the 3D domain, it is still unexplored. One of the first works is introduced by Azuma \etal~\cite{azuma2022scanqa}. Based on ScanNet~\cite{dai2017scannet} scenes and ScanRefer~\cite{chen2020scanrefer} object descriptions, the authors create the ScanQA dataset with the additional task of grounding the target object(s) with a bounding box in the scenes. 
Their baseline architecture consists of two encoder branches, one of which utilizes a PointNet++~\cite{qi2017pointnet++} backbone to encode the scene into object proposals. A transformer based fusion module~\cite{yu2019mcan} produces a final vector from which the answer is predicted. 
Furthermore, Prelli \etal achieve the current state-of-the-art on the ScanQA~\cite{azuma2022scanqa} benchmark with their method where they apply knowledge from the 2D domain into the 3D domain. They pretrain a CLIP~\cite{radford2021learning} encoder to align the scene features with the image and question embeddings. In the next training step, the CLIP module is used to encode the question sequence. Similarly, the answer is predicted from the final $<$end$>$ token representation of the question.
In contrast to previous works, we solely train on the 3D scenes with SoftGroup~\cite{vu2022softgroup} as a backbone. Apart from a transformer encoder, we also implement a transformer decoder to generate rather than predict the answer. 


\mypara{Reinforcement Learning for Sequence Generation.} Rennie \etal~\cite{rennie2017selfcritical} introduce reinforcement learning to the task of image captioning where they utilize the REINFORCE algorithm~\cite{Williams2004SimpleSG} and optimize their model directly on the non differentiable CIDEr~\cite{vedantam2015cider} metric. Their LSTM \cite{lstm} model is seen as the "agent" that interacts with the "environment", which is the words and image features. The network acts as the "policy" that determines the "action" taken by the agent, which in this case is the prediction of the next word. Following an action, the model updates its "state", i.e. weights. Once a sentence is generated, the model receives a "reward" in form of the CIDEr score of the generated sentence. 
Vedantam \etal~\cite{cornia2020meshedmemory} apply the same training method on their transformer based architecture with a small variation, where they generate $k$ sentences with the beam search algorithm and baseline each sentence on the average reward of all sentences.
Luo \etal~\cite{luo2020better} introduce a better variant of the original self-critical sequence training (SCST~\cite{rennie2017selfcritical}) where they sample $k$ sentences (using random sampling) and calculate for each sentence the average reward of the rest as a baseline.

\section{Method}
\begin{figure*}[t]
  \centering
   \includegraphics[width=0.9\linewidth]{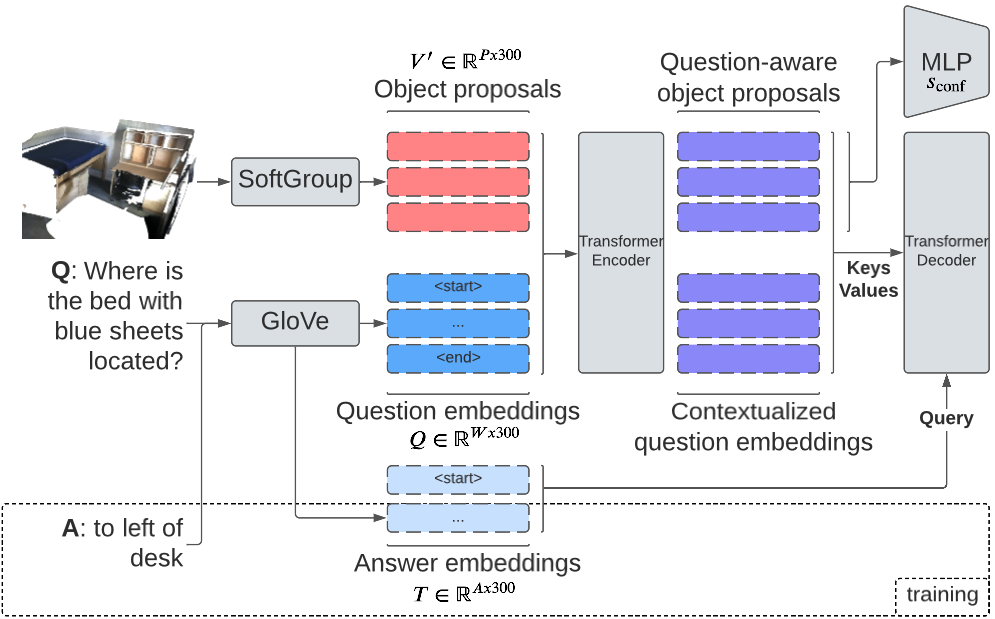}
   \caption{Overview of our model architecture. The input scene is encoded into object proposals $V'$ with SoftGroup~\cite{vu2022softgroup}. The question and answer tokens are turned into word embeddings ($Q$ and $T$) with GloVe~\cite{pennington2014glove}. The question sequence and object proposals are concatenated as a single sequence and fed into the transformer encoder. The contextualized sequence is then forwarded to a transformer decoder as keys and values, while the embedded answer sequence is passed as query during training with XE loss. During inference, only the $<$start$>$ token embedding is used as a query in the decoder to begin the sentence. Best viewed in color.}
   \label{fig:model}
\end{figure*}

In this section, we explain our model architecture (Figure \ref{fig:model}) and training method. The input of our model is a 3D point cloud with RGB and normals features. The second input is the question sequence, and the output is a token sequence of the generated answer. Overall, our model can be divided into 3 segments: SoftGroup, transformer encoder-decoder, and object localization. Our training method consists of 3 stages, which we will explain in detail in Section \ref{sec:training}.

\subsection{Model}
\label{sec:model}

\mypara{SoftGroup.} Instead of relying on a pointwise method for the backbone network~\cite{qi2019deep} like in previous works~\cite{chen2020scanrefer, chen2021scan2cap, azuma2022scanqa, parelli2023clipguided}, we employ the 3D-sparse-convolution based method SoftGroup~\cite{vu2022softgroup} to extract denser semantic information of the object proposals. This method has shown better performance and speed on the object detection task~\cite{vu2022softgroup}. In addition, the instance masks of the generated object proposals offer a unique identity for objects and thus provide better scene understandings for the answers.  With that, we generate $P$ semantically rich object proposals $V \in \mathbb{R}^{P\times32}$ from the point cloud scene. To match the dimension of our question token embeddings, we train a linear layer that expands the object proposals to $V' \in \mathbb{R}^{P\times300}$.

\mypara{Transformer Encoder-Decoder.} We encode our $W$ question tokens with GloVe~\cite{pennington2014glove} embeddings as $Q \in \mathbb{R}^{W\times300}$. Then, we add positional encodings to both of our modalities representations. For the question embeddings, we follow the original transformer positional encoding~\cite{vaswani2017attention}. As for the object proposals, we add the normalized center points $Z \in \mathbb{R}^{P\times3}$ to the last 3 dimensions of each (expanded) object proposal. Transformers have shown great performance when it comes to sequence-to-sequence generation. There are several approaches to encoding two sequences of different modalities with transformers. Following the approaches mentioned in the survey by Xu \etal~\cite{xu2022multimodal}, we choose early concatenation, which enables the model to equally encode the scene information into the question and the question information into the scene. This method has shown to well preserve the global multi-modal context~\cite{sun2019videobert, lin2021interbert, xu2022multimodal}, which is necessary for both answer generation and object localization tasks. Hence, we concatenate both sequences into one sequence $S \in \mathbb{R}^{L\times300}$ where $L = P + W$ and feed it into a two-layer transformer encoder. The sequence $S$ acts as the keys, values and query and is encoded into one sequence $S' \in \mathbb{R}^{L\times300}$ containing the contextualized object proposals and question embeddings.
The contextualized sequence is fed into a two-layer transformer decoder as keys and values. The target sequence containing GloVe word embeddings of the answer $T \in \mathbb{R}^{A\times300}$ is used as the query.

\mypara{Object Localization.} After encoding the full multi-modal sequence, we feed the question-aware object proposals into an MLP to predict their confidence scores $s_\text{conf} \in \mathbb{R}^{P\times1}$. The object proposal with the highest confidence score is considered as our target object.

\subsection{Training}
\label{sec:training}

First, we pretrain SoftGroup~\cite{vu2022softgroup} with the ScanRefer~\cite{chen2020scanrefer} object classes. We experiment with different input features and find that RGB + normals features result in the best overall scores. Our object detection scores from SoftGroup~\cite{vu2022softgroup} can be found in the appendix. Since the forward pass of SoftGroup~\cite{vu2022softgroup} is significantly more time-consuming than the forward pass of our language model, we precompute the object proposals from the pretrained SoftGroup~\cite{vu2022softgroup} model and save them on disk before using them for our visual-language model. During prediction on the test sets, we re-activate SoftGroup~\cite{vu2022softgroup} and do the full forward pass. SoftGroup~\cite{vu2022softgroup} is trained end-to-end on a multitask loss $L_\text{softgroup}$, which encompasses the total loss for the first training stage.

Next, we train our question answering model on word level cross entropy (XE) loss:
$L_\text{ans} = - \sum_{t \in T}\sum_{z \in Z} y_{t,z} log(\hat{y}_{t,z})$
where \(T\) is the ground truth answer including the $<$end$>$ token and \(Z\) is the training vocabulary. \(y_{t,z}\) has the value \(1\) when the current ground truth token \(t\) matches the vocabulary token \(z\) and \(0\) otherwise. \(\hat{y}_{t,z}\) is the predicted probability of the token \(z\) in the Softmax output for the word at step \(t\). The model is trained with the teacher forcing scheme, where we pass the ground truth previous words as the query to predict the next word at each time step. Simultaneously, we train our object localization branch on cross entropy loss $L_\text{loc}$, similar to~\cite{chen2020scanrefer, azuma2022scanqa}. Thus, our total loss for the second training stage is $L = L_\text{ans} + L_\text{loc}$.

After our question-answering model converges on the CIDEr score accuracy, we drop the word level XE loss $L_\text{ans}$ and switch to reinforcement learning, while keeping the object localization loss $L_\text{loc}$. Here, we apply the self-critical sequence training~\cite{rennie2017selfcritical} method and train directly on the CIDEr score. We treat our transformer model as the "agent", the question \& answer words and object proposals as the "environment", our network parameters as the "policy" $p_{\theta}$, the prediction of the next word as the "action", and the CIDEr score of the generated answer as the "reward". Instead of sampling the answer sequence like in~\cite{rennie2017selfcritical}, we generate it using test-time greedy decoding to get $w^g$ where $w^g = (w^g_1, ..., w^g_T)$ and $w^g_t$ is the word with the maximum likelihood at time step $t$. Our loss can be expressed as the negative expected reward:
\begin{equation}
    L_\text{cider}(\theta) = - \mathbb{E}_{w^g\sim p_{\theta}}[r(w^g)]
    \label{eq:cider_loss}
\end{equation} 
where $r(.)$ is the reward function (CIDEr score). As for the baseline, we generate $k$ answers using beam search decoding. We keep track of the top-$k$ answers and predict the next word until we reach the $<$end$>$ token for all top $k$ sequences. We take the average reward of the $k$ answers as the baseline reward $r_\text{VQA}^{b}$. In addition to the reward from the generated answer, we also train a Visual Question Generation (VQG) module by simply switching the input (question) and output (answer) of our transformer model. Since we treat the VQA task as a sequence generation problem, our model can be easily switched to the inverse task. Once we generate an answer from the VQA module, we feed it into the frozen VQG module to greedily generate a question $q^g$. The same thing is also done with the generated baseline answers, which results in $k$ baseline questions. Similar to VQA, we get the CIDEr scores for the generated question $r_\text{VQG}^{g}$ and for the baseline questions $r_\text{VQG}^{b}$.
With that, we can express the gradient of our loss as:
\begin{equation}
    \resizebox{.9\hsize}{!}{$
        \nabla_{\theta}L_\text{cider}(\theta) = - ((r_\text{VQA}^{g} - r_\text{VQA}^{b}) + (r_\text{VQG}^{g} - r_\text{VQG}^{b})) \nabla_{\theta} log p_{\theta}(w^g)
    $}
    \label{eq:cider_loss_gradient}
\end{equation}
where $r_\text{VQA}^g$ is the reward for the generated answer $w^g$. During our experiments, we find that using greedy decoding for the VQA baseline, as in ~\cite{rennie2017selfcritical}, does not yield any improvement, since the sampled sentence in our case usually has a worse CIDEr score than the answer generated with greedy decoding. Thus, we experiment with greedy decoding for generating the answer and sampling for the baseline. However, we also find that sentences generated with beam search have worse results than the ones generated greedily. Therefore, we conduct experiments with using beam search as our baseline and see a noticeable improvement. When using a beam size of 2, the reward difference between the generated answer and the average of the baseline answers becomes too small and crashes the accuracies after few epochs. Increasing the beam size to 3 widens the difference in rewards and stabilizes our training. The final total loss for the third training stage is $L = L_\text{cider} + L_\text{loc}$.

\subsection{Inference}

During inference, we re-activate SoftGroup~\cite{vu2022softgroup} to generate object proposals. As for the transformer decoder, we apply greedy decoding to generate the answer sequence beginning with the token $<$start$>$. Once we reach the $<$end$>$ token, our decoder stops. We determine the confidence score for each object proposal with the object localization branch and pick the one with the highest value as our target object. The object class of the target object is determined by the classification branch of SoftGroup~\cite{vu2022softgroup}.

\section{Experiments}

\begin{table*}[t]
  \centering
    {
        \begin{tabular}{|l|c|c|c|c|c|c|c|}
         \hline
        Model & BLEU-1 & BLEU-4 & ROUGE & METEOR & CIDEr \\
         \hline\hline
        \textbf{Test w/ object IDs} & & & & & \\
         ScanQA~\cite{azuma2022scanqa} & 31.56 & 12.04 & 34.34 & 13.55 & 67.29 \\
         CLIP-guided~\cite{parelli2023clipguided} & 32.72 & \textbf{14.64} & 35.15 & 13.94 & 69.53 \\
         Gen3DQA (XE loss) & 35.24 & 10.79 & 33.50 & 13.61 & 64.83 \\
         Gen3DQA & \textbf{39.30} & 12.24 & \textbf{35.78} & \textbf{14.99} & \textbf{72.22} \\
        \textbf{Test w/o object IDs} & & & & & \\
         ScanQA~\cite{azuma2022scanqa} & 30.68 & 10.75 & 31.09 & 12.59 & 60.24 \\
         CLIP-guided~\cite{parelli2023clipguided} & 32.70 & \textbf{11.73} & 32.41 & 13.28 & 62.83 \\
         Gen3DQA (XE loss) & 35.08 & 10.62 & 30.99 & 12.87 & 60.05 \\
         Gen3DQA & \textbf{38.07} & 11.61 & \textbf{33.03} & \textbf{14.28} & \textbf{66.57} \\
         \hline
      \end{tabular}
  }
  \break
  \caption{Image captioning metrics scores of previous methods and ours on the ScanQA~\cite{azuma2022scanqa} test benchmark with and without object IDs. At the time of evaluation on the benchmark website, the SPICE~\cite{spice2016} score is not available.}
  \label{tab:scores}
\end{table*}

\begin{figure*}[t]
  \centering
   \includegraphics[width=0.8\linewidth]{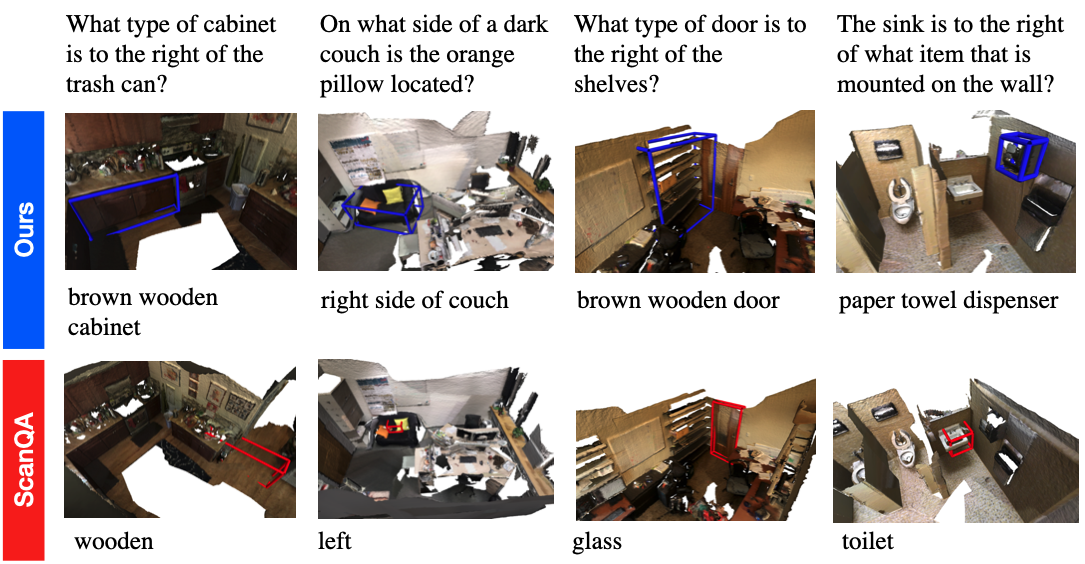}
   \break
   \includegraphics[width=0.8\linewidth]{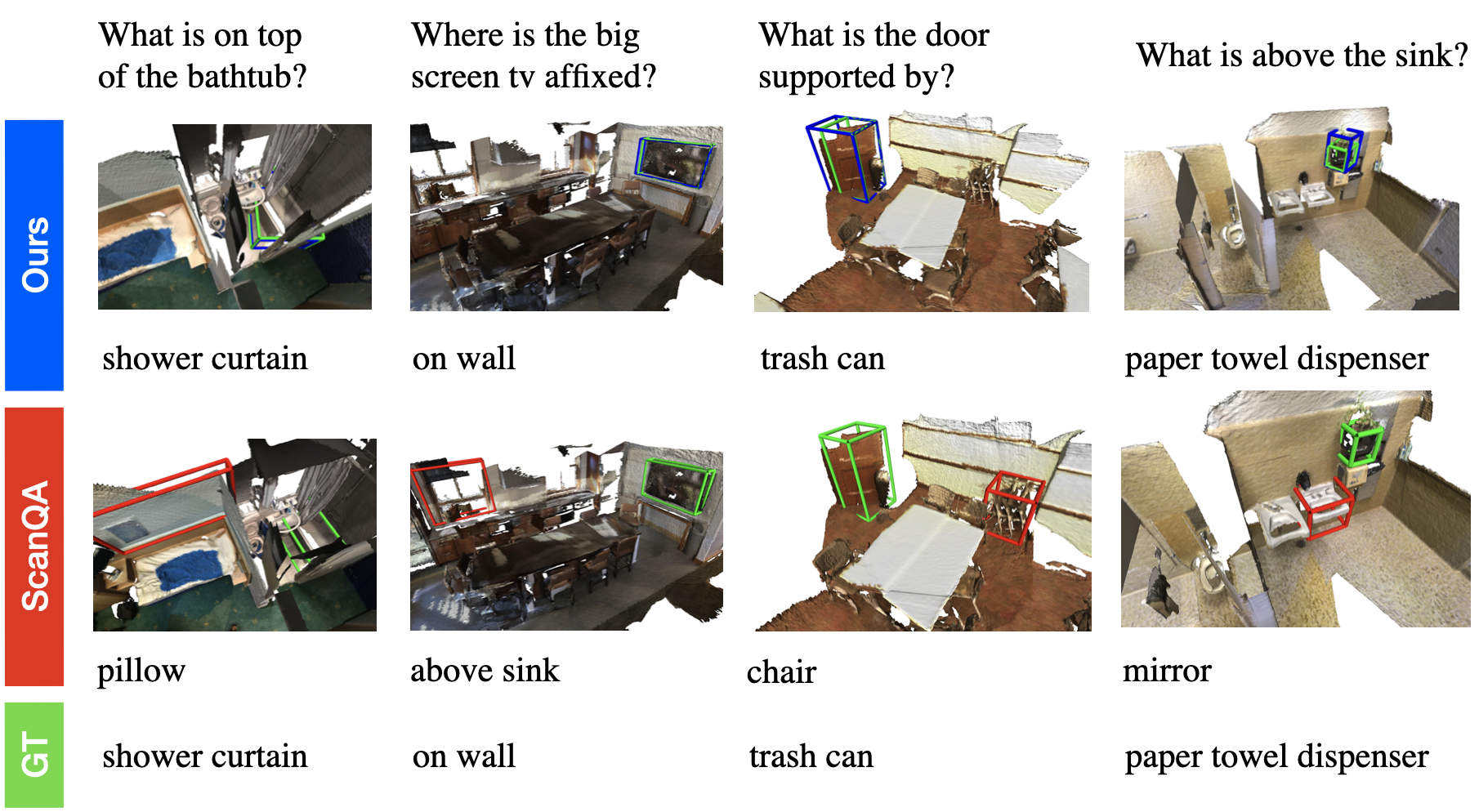}
   \caption{Example questions and answers from the test set with object IDs (top) and the validation set (bottom). We compare the results of our model (blue) to ScanQA~\cite{azuma2022scanqa} (red) and the ground truth (GT) (green). Below every image is the predicted or generated answer. Since we do not axis-align our scenes, the bounding boxes in our model look tilted. We include more examples in the appendix. Best viewed in color.}
   \label{fig:examples_test}
\end{figure*}


\subsection{Data}

We train and test our model on the ScanQA~\cite{azuma2022scanqa} dataset. The 3D scenes are from the ScanNet~\cite{dai2017scannet} dataset, while the questions are based on the ScanRefer~\cite{chen2020scanrefer} object descriptions. Hence, the categories of the objects in question are from the ScanRefer~\cite{chen2020scanrefer} classes. Moreover, we treat questions with multiple answers as multiple training samples, where every sample contains the same question and one of the answers. This introduces 952 additional training samples. Furthermore, we evaluate our model on both test sets of ScanQA~\cite{azuma2022scanqa} on the image captioning metrics BLUE-1~\cite{BLEU}, BLEU-4~\cite{BLEU}, ROUGE~\cite{lin-2004-rouge}, METEOR~\cite{banerjee-lavie-2005-meteor} and CIDEr~\cite{vedantam2015cider} and exclude the EM@1 and EM@10 accuracies since we do not have answer classification in our model. Our scores are calculated by uploading our question-answering results to the ScanQA~\cite{azuma2022scanqa} benchmark server \footnote{\url{https://eval.ai/web/challenges/challenge-page/1715/overview}}, where at the time of writing the SPICE~\cite{spice2016} score is returning with the value of 0.0 and is thus not included.

\subsection{Implementation Details}

We follow the implementation of MINSU3D \footnote{\url{https://github.com/3dlg-hcvc/minsu3d}} for training SoftGroup~\cite{vu2022softgroup} and change the class mappings in the data preparation phase to fit the ScanRefer~\cite{chen2020scanrefer} object classes. We use Adam~\cite{kingma2017adam} optimizer for training our language model with a learning rate of 8e-5 when training on XE loss and 2e-5 when training with reinforcement learning. In both cases, we apply cosine annealing~\cite{loshchilov2017sgdr} for scheduling. We train with a batch size of 64 on a GeForce RTX 2080 Ti. Our models are implemented in PyTorch~\cite{paszke2017automatic}. For data augmentation, we randomly replace one random word in the question with the $<$unk$>$ token. When training on XE loss, our model converges on the CIDEr score after 80,000 iterations. As for training with the REINFORCE algorithm, our best model converges after 100,000 iterations.

\subsection{Quantitative Analysis}

We show in Table \ref{tab:scores} our final results in comparison to ScanQA~\cite{azuma2022scanqa} and CLIP-guided~\cite{parelli2023clipguided}. Our method outperforms previous works on the conditional image captioning metrics and especially on the more challenging CIDEr score. Unlike the CLIP-based method~\cite{parelli2023clipguided}, our model only requires 3D point cloud data to train. By training directly on the CIDEr score, our model performance is significantly improved on the rest of the metrics too.

\begin{table}
  \centering
  {
      \begin{tabular}{|l|c|c|c|}
        \hline
        & Acc@0.5 \\
        \hline
        ScanQA~\cite{azuma2022scanqa} & 15.42 \\
        CLIP-guided~\cite{parelli2023clipguided} & 21.22 \\
        Gen3DQA & \textbf{23.79} \\
        \hline
      \end{tabular}
  }
  \break
  \caption{Object localization accuracy Acc@0.5 of previous methods and ours on the ScanQA~\cite{azuma2022scanqa} validation set.}
  \label{tab:scores_loc}
\end{table}

Furthermore, we also look at the object localization task on the validation set in comparison to previous methods (Table~\ref{tab:scores_loc}). Even though the current state-of-the-art trains with additional image data, our model achieves a noticeable improvement on the Acc@0.5 accuracy. We also see that our early concatenation method is superior to the fusion module of ScanQA~\cite{azuma2022scanqa} in multi-modal context understanding. With that, our method presents a stronger understanding of the scene and question and can generate context-aware answers naturally while localizing relevant objects significantly better than previous methods.

\subsection{Qualitative Analysis}

In Figure~\ref{fig:examples_test} we showcase samples from the test set (with object IDs) where our model generates better answers than ScanQA~\cite{azuma2022scanqa} predicts, while localizing a meaningful target object. Overall, we see that our model generates longer answers that contain more information. In fact, compared to ScanQA~\cite{azuma2022scanqa}, the average number of words in an answer from our model is 1.87/1.92 (test set with and without object IDs) compared to ScanQA~\cite{azuma2022scanqa} with 1.41/1.47. Furthermore, we show in Figure~\ref{fig:examples_test} samples from the validation set where our model performs better object localization than ScanQA~\cite{azuma2022scanqa} while also generating the correct answer. We see in the samples that our model performs well when the question requires spatial awareness and can also extract details about object types and looks.

\begin{table*}[t]
  \centering
  {
      \begin{tabular}{|l|c|c|c|c|c|c|c|}
        \hline
        Model & BLEU-1 & BLEU-4 & ROUGE & METEOR & CIDEr \\
        \hline\hline
         Gen3DQA (single object) & 35.4 & \textbf{10.52} & \textbf{33.39} & 13.62 & \textbf{64.91} \\
         Gen3DQA (multiple objects) & \textbf{36.02} & 10.21 & 32.84 & \textbf{13.68} & 64.51 \\
         Gen3DQA (w/o object localization) & 34.29 & 10.02 & 31.2 & 12.99 & 59.35 \\
        \hline
      \end{tabular}
  }
  \break
  \caption{Scores of our model trained on XE loss once with targeting a single object, once multiple objects, and once without object localization at all. Evaluation is done on the validation set.}
  \label{tab:scores_ablation_multiobj}
\end{table*}


\begin{table*}
  \centering
  {
      \begin{tabular}{|l|c|c|c|c|c|c|c|}
        \hline
        Model & BLEU-1 & BLEU-4 & ROUGE & METEOR & CIDEr \\
        \hline\hline
        \textbf{Valid} & & & & & \\
         Gen3DQA (w/o VQG reward) & 39.12 & \textbf{13.2} & 35.48 & 14.89 & 71.39 \\
         Gen3DQA (w/ VQG reward) & \textbf{39.53} & 12.7 & \textbf{35.97} & \textbf{15.11} & \textbf{71.97}  \\
        \textbf{Test w/ object IDs} & & & & & \\
         Gen3DQA (w/o VQG reward) & 38.89 & \textbf{12.67} & 35.35 & 14.82 & 71.09 \\
         Gen3DQA (w/ VQG reward) & \textbf{39.30} & 12.24 & \textbf{35.78} & \textbf{14.99} & \textbf{72.22} \\
        \textbf{Test w/o object IDs} & & & & & \\
         Gen3DQA (w/o VQG reward) & 37.61 & \textbf{12.00} & 32.57 & 14.09 & 65.58 \\
         Gen3DQA (w/ VQG reward) & \textbf{38.07} & 11.61 & \textbf{33.03} & \textbf{14.28} & \textbf{66.57} \\
        \hline
      \end{tabular}
  }
  \break
  \caption{Scores of our model trained with reinforcement learning with and without the additional reward of Visual Question Generation (VQG).}
  \label{tab:scores_ablation_vqg}
\end{table*}

\subsection{Ablation Studies}


\mypara{Does multi-object localization help?} In ScanQA~\cite{azuma2022scanqa} the authors experiment with training the object localization branch on binary cross entropy (BCE) loss. This enables the model to decide for each object whether it should be targeted or not, allowing multiple objects to be selected. 
Overall, we don't see a clear performance improvement in our model from targeting multiple objects (Table~\ref{tab:scores_ablation_multiobj}). We also conduct an experiment where we train our model without the object localization loss and see that by training our model to localize the target object, it becomes better at generating answers.

\mypara{Does VQG reward improve answer generation?} We hypothesize in the beginning that by generating a better answer, it becomes easier to regenerate the original question from it. Hence, we train a question-generation module (VQG) and use it to generate a question from our generated answer during reinforcement learning. We then add the CIDEr score of the generated question as an additional reward. The results in Table~\ref{tab:scores_ablation_vqg} show that training with the additional question generation reward yields better answer generation scores.

\section{Conclusion and Future Work}

In this work, we propose a new architecture for the task of 3D visual question answering to generate free-form answers.
We directly train our model on the CIDEr metric using a version of the REINFORCE algorithm~\cite{Williams2004SimpleSG, rennie2017selfcritical}. In addition, we introduce the inverse task of question generation to enhance our question-answering model during reinforcement learning. Our experiments and results show that our method outperforms the current state-of-the-art on the image captioning metrics of the ScanQA~\cite{azuma2022scanqa} benchmark. For future work, we encourage the research community to further explore the dual tasks of question answering and generation. For instance, we suggest jointly training both tasks without freezing any weights. We also look forward to future works to explore and develop better answer generation models instead of answer classification ones for questions in 3D environments.

\section{Acknowledgement}

This work was supported by the ERC Starting Grant Scan2CAD (804724) and the German Research Foundation (DFG)
Research Unit “Learning and Simulation in Visual Computing”.

{\small
\bibliographystyle{ieee_fullname}
\bibliography{egbib}
}

\clearpage
\appendix

\section*{Supplementary Material}

\counterwithin{figure}{section}
\counterwithin{table}{section}
\counterwithin{equation}{section}

This supplementary material provides additional experiment results and evaluations, such as the performance of the SoftGroup~\cite{vu2022softgroup} backbone trained on ScanRefer~\cite{chen2020scanrefer} classes (Section \ref{sec:additional_quantitative_analysis_softgroup}). We also include the question-answering scores on the different types of questions in comparison to ScanQA~\cite{azuma2022scanqa} (Section \ref{sec:additional_quantitative_analysis_question_types}). Apart from that, we show additional ablation studies in Section \ref{sec:additional_ablation_studies} and further qualitative analysis results in Section \ref{sec:additional_qualitative_analysis}.

\section{Additional Quantitative Analysis Results}

\subsection{SoftGroup Trained on ScanRefer Classes}
\label{sec:additional_quantitative_analysis_softgroup}

\begin{table*}
    \begin{center}
        \resizebox{\textwidth}{!} {
            \begin{tabular}{|l|c|c|c|c|c|c|c|c|}
                \hline
                    Point cloud features & AP & AP 50\% & AP 25\% & Bbox AP 50\% & Bbox AP 25\% & AR & RC 50\% & RC 25\% \\
                    \hline\hline
                    xyz & 40.4 & 60.6 & 72.1 & 54.3 & 66.8 &	49.8 & 72.1 & 83.6 \\
                    xyz + rgb & 40.6 &	60.9 &	74.2 &	53.5 &	68.1 &	49.7 &	71.6 &	\textbf{84.3} \\
                    xyz + rgb + normals & \textbf{42.0} &	\textbf{62.2} &	\textbf{74.5} &	\textbf{57.1} &	\textbf{69.3} &	\textbf{51.3} &	\textbf{73.6} &	83.8 \\
                    \hline
            \end{tabular}
        }
    \end{center}
\caption{Evaluation scores of SoftGroup~\cite{vu2022softgroup} trained with ScanRefer~\cite{chen2020scanrefer} object classes. We report our scores on the ScanNet~\cite{dai2017scannet} validation set.}
\label{tab:softgroup_scores}
\end{table*}

We show our evaluation results (Table \ref{tab:softgroup_scores}) of SoftGroup~\cite{vu2022softgroup} trained on ScanNet~\cite{dai2017scannet} scenes with different input features with ScanRefer~\cite{chen2020scanrefer} object classes. We see that having RGB and normals features yields the best overall scores.

\subsection{Question Types}
\label{sec:additional_quantitative_analysis_question_types}

\begin{table*}[t]
  \centering
    {
      \begin{tabular}{|l|c|c|c|c|c|c|c|}
        \hline
        Model & BLEU-1 & BLEU-4 & ROUGE & METEOR & CIDEr \\
         \hline\hline
        \textbf{Object} & & & & & \\
         ScanQA \cite{azuma2022scanqa} & 23.94 & 0.00 & \textbf{50.05} & 10.62 & 26.01 \\
         Gen3DQA & \textbf{27.27} & 0.00 & 27.23 & \textbf{11.97} & \textbf{55.13} \\
        \textbf{Color} & & & & & \\
         ScanQA \cite{azuma2022scanqa} & 43.92 & 0.00 & \textbf{84.42} & 22.61 & 47.68 \\
         Gen3DQA & \textbf{45.76} & 0.00 & 48.77 & \textbf{22.92} & \textbf{83.22} \\
        \textbf{Object Nature} & & & & & \\
         ScanQA \cite{azuma2022scanqa} & \textbf{41.65} & 0.00 & \textbf{73.26} & 16.54 & 41.61 \\
         Gen3DQA & 41.63 & 0.00 & 39.51 & \textbf{17.61} & \textbf{73.72} \\
        \textbf{Place} & & & & & \\
         ScanQA \cite{azuma2022scanqa} & 28.78 & 9.55 & \textbf{57.00} & 11.49 & 28.19 \\
         Gen3DQA & \textbf{43.11} & \textbf{12.32} & 38.32 & \textbf{14.81} & \textbf{72.74} \\
        \textbf{Number} & & & & & \\
         ScanQA \cite{azuma2022scanqa} & 44.29 & 0.00 & \textbf{72.15} & 19.16 & 46.05 \\
         Gen3DQA & \textbf{51.97} & \textbf{0.04} & 50.18 & \textbf{20.99} & \textbf{74.93} \\
        \textbf{Other} & & & & & \\
         ScanQA \cite{azuma2022scanqa} & 22.26 & 0.00 & \textbf{45.39} & 9.96 & 26.30 \\
         Gen3DQA & \textbf{37.52} & \textbf{16.77} & 30.40 & \textbf{14.78} & \textbf{64.11} \\
        \textbf{Total} & & & & & \\
         ScanQA \cite{azuma2022scanqa} & 29.47 & 9.55 & 32.37 & 12.60 & 61.66 \\
         Gen3DQA & \textbf{39.53} & \textbf{12.70} & \textbf{35.97} & \textbf{15.11} & \textbf{71.97} \\
        \hline
      \end{tabular}
    }
    \break
  \caption{Image captioning metrics scores for different types of questions in the ScanQA~\cite{azuma2022scanqa} validation set.}
  \label{tab:scores_types}
\end{table*}

We compare our results with the ScanQA~\cite{azuma2022scanqa} baseline on the different types of questions in the validation set (Table \ref{tab:scores_types}). Since the question types split is not publicly available, we split the validation set based on the beginning words of every question, as mentioned in the ScanQA~\cite{azuma2022scanqa} paper. With that, we get the same number of questions as ScanQA~\cite{azuma2022scanqa} for each type. Overall, our model outperforms the baseline in all question types on all image captioning metrics except ROUGE~\cite{lin-2004-rouge}. The biggest difference in scores can be observed in the "other" category, where our model has a BLEU-4 score of 16.77 compared to 0.00 of the baseline.

\section{Additional Ablation Studies}
\label{sec:additional_ablation_studies}

\begin{table*}[ht]
  \centering
  {
      \begin{tabular}{|l|c|c|c|c|c|c|c|}
        \hline
        Model & BLEU-1 & BLEU-4 & ROUGE & METEOR & CIDEr \\
        \hline\hline
         Gen3DQA (w/o target embeddings) & \textbf{35.4} & 10.52 & \textbf{33.39} & \textbf{13.62} & \textbf{64.91} \\
         Gen3DQA (w/ target embeddings) & 34.65 & \textbf{11.07} & 33.31 & 13.57 & 64.71 \\
        \hline
      \end{tabular}
  }
  \break
  \caption{Image captioning metrics scores of our model trained on XE loss once with and once without target embeddings. Evaluation is done on the validation set.}
  \label{tab:scores_ablation_tgtembed}
\end{table*}

\textbf{Do target embeddings help?} Our aim in this experiment is to pass a signal from our object localization branch to the decoder by adding information about the target object proposal. Therefore, we train 0 \& 1 embeddings and add the 1 embedding vector to the encoded object proposal with the highest confidence score and the 0 embedding vector to the rest. Our results in Table \ref{tab:scores_ablation_tgtembed} show that there is no significant improvement when using the target embeddings. We assume the reason is the low object localization accuracy of our model (23.79 on Acc@0.5), because of which it does not get an accurate signal most of the time.

\begin{table*}[ht]
  \centering
    \begin{tabular}{|l|c|c|c|c|c|c|c|}
     \hline
    Model & BLEU-1 & BLEU-4 & ROUGE & METEOR & CIDEr \\
     \hline\hline
     Gen3DQA (SCST switched) & 38.25 & 13.01 & 35.36 & 14.82 & 70.96 \\
     Gen3DQA (w/o VQG) & \textbf{39.12} & \textbf{13.2} & \textbf{35.48} & \textbf{14.89} & \textbf{71.39} \\
     \hline
  \end{tabular}
  \break
  \caption{
    Experiment results on the validation set. Models are trained without VQG reward.
  }
  \label{tab:scst}
\end{table*}

\textbf{Does using beam search as a basesline for SCST help?} In the SCST paper the authors use the greedy decoding output for the baseline reward. In our case, the sampled sentences are almost always worse than the ones generated by greedy decoding. As our model tries to make the reward gap positive, it becomes much worse after 5 epochs, where the CIDEr score goes below 22. Therefore, we experiment with switching the sampled sentence and the greedily generated one and report our findings in Table \ref{tab:scst} (Gen3DQA (SCST switched)). As can be seen, using beam search for the baseline reward performs better.

\section{Additional Qualitative Analysis Results}
\label{sec:additional_qualitative_analysis}

\begin{figure*}[htbp]
  \centering
   \includegraphics[width=1.0\linewidth]{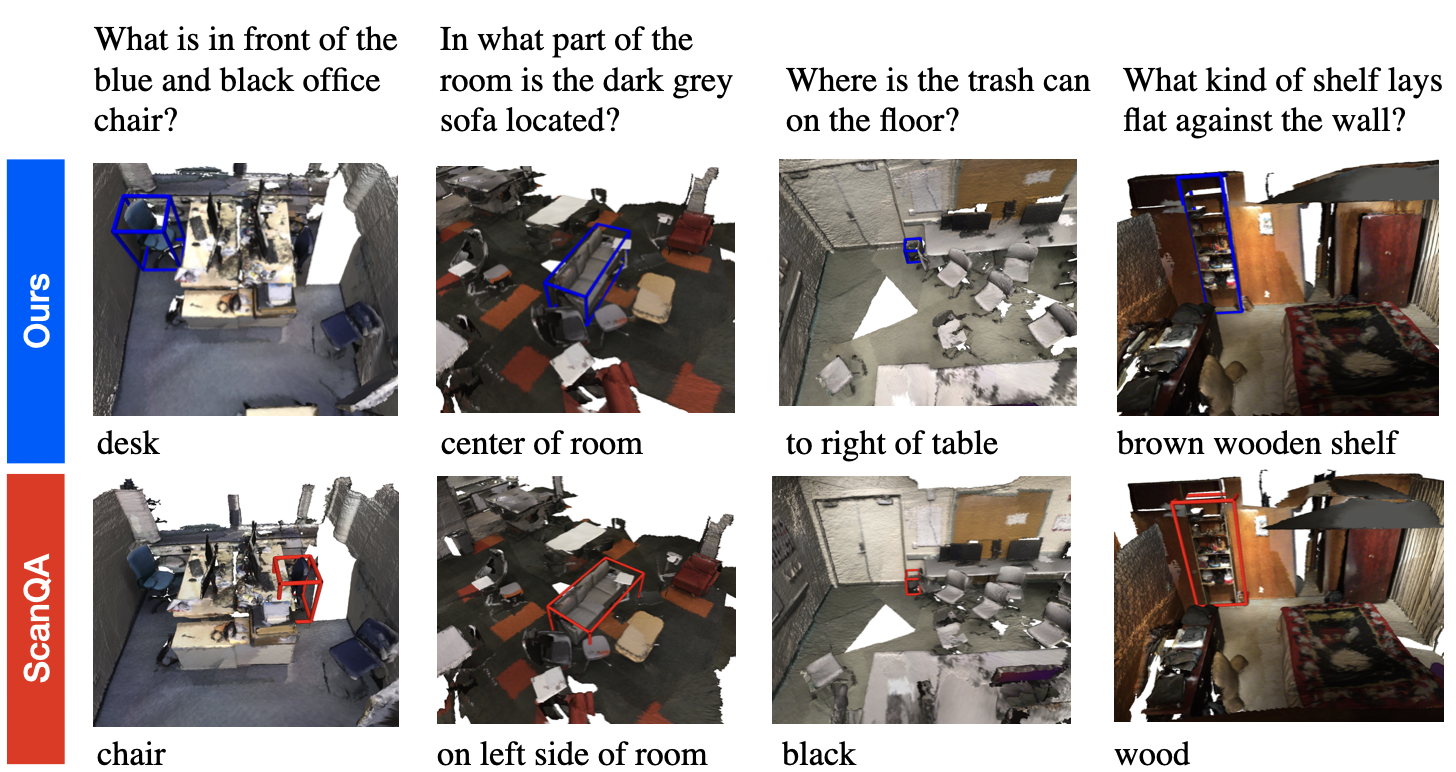}
   \break
   \includegraphics[width=1.0\linewidth]{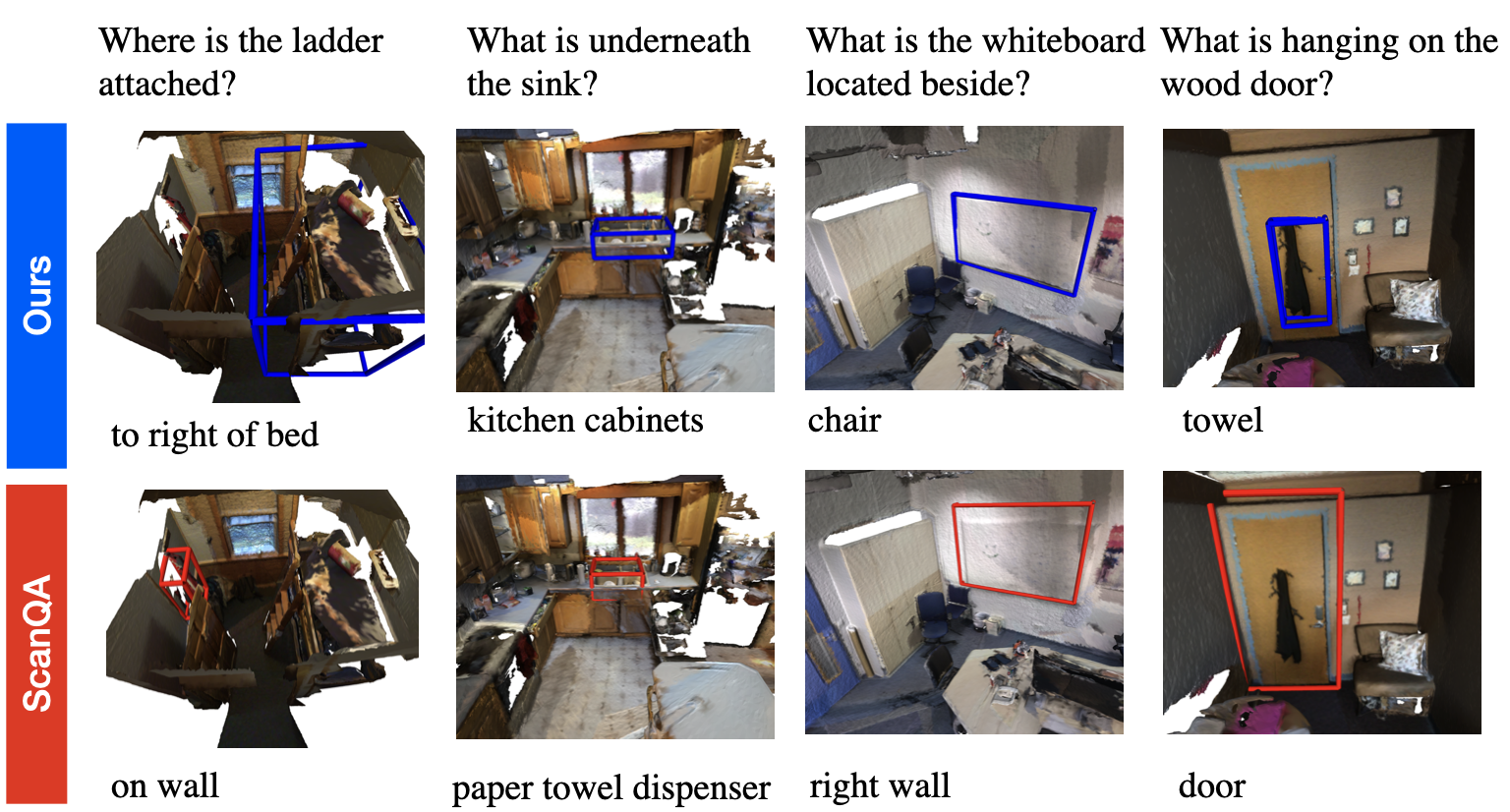}
   \caption{Example questions and answers from the test set without object IDs. We compare the results of our model (blue) to ScanQA~\cite{azuma2022scanqa} (red). Below every image is the predicted or generated answer. Since we do not axis-align our scenes, the bounding boxes in our model look tilted. Best viewed in color.}
   \label{fig:examples_test_wo_obj}
\end{figure*}

\begin{figure*}[ht]
  \centering
   \includegraphics[width=1.0\linewidth]{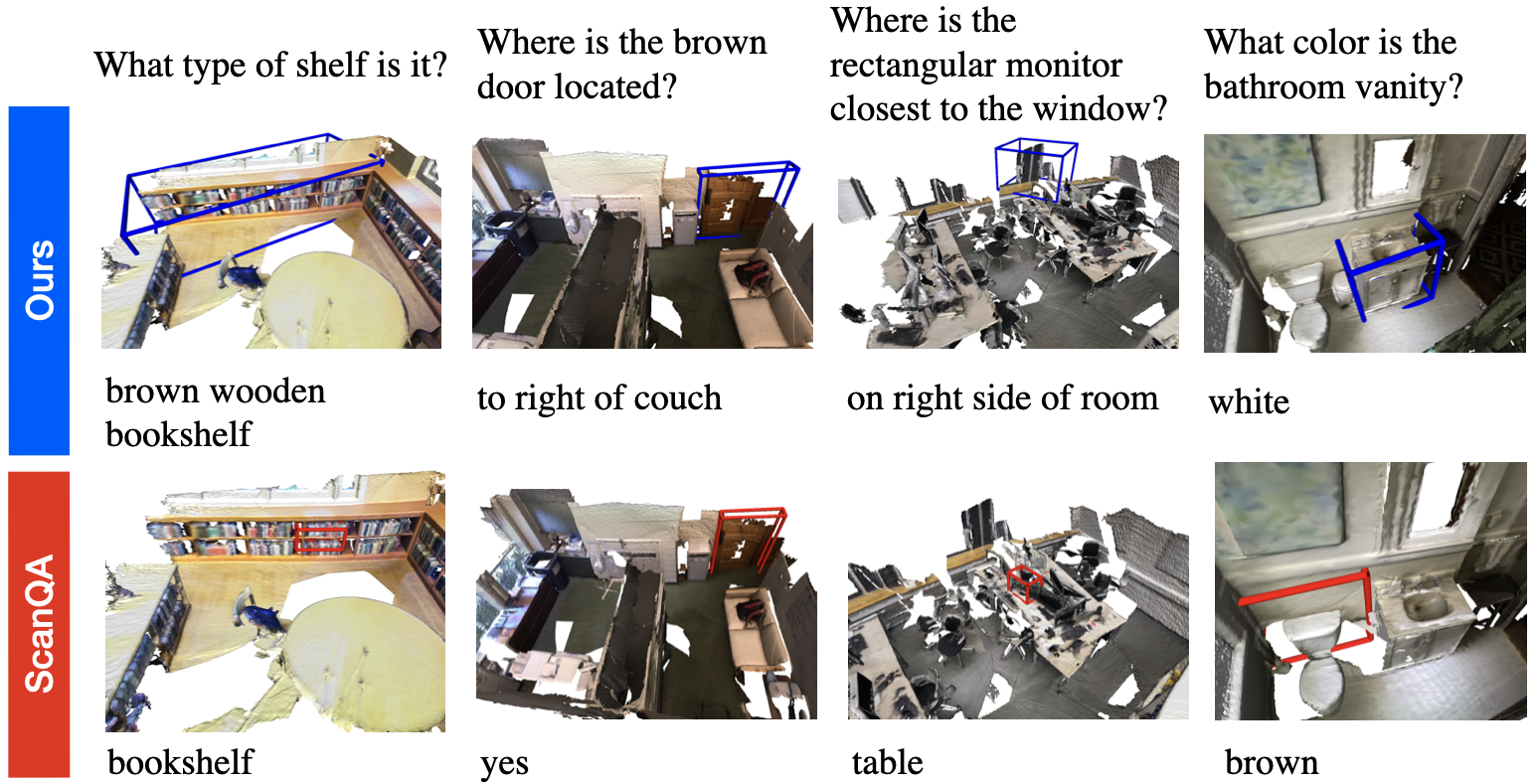}
   \caption{Example questions and answers from the test set with object IDs. We compare the results of our model (blue) to ScanQA~\cite{azuma2022scanqa} (red). Below every image is the predicted or generated answer. Since we do not axis-align our scenes, the bounding boxes in our model look tilted. Best viewed in color.}
   \label{fig:examples_test_w_obj}
\end{figure*}

In Figures \ref{fig:examples_test_wo_obj} and \ref{fig:examples_test_w_obj} we show additional examples of our model compared to ScanQA~\cite{azuma2022scanqa}. We see that while our model localizes meaningful targets, it generates longer and/or better answers than ScanQA~\cite{azuma2022scanqa}.

\end{document}